\documentclass[runningheads, envcountsame,a4paper]{llncs}
\usepackage[pdftex]{graphicx}
\usepackage{epstopdf}
\usepackage[misc]{ifsym}
\usepackage{array}
\usepackage{multirow}
\usepackage{enumitem}
\usepackage{multicol}
\usepackage{booktabs}
\usepackage[caption=false]{subfig}
\usepackage{bm}
\usepackage{makecell}
\usepackage{xcolor}
\usepackage{verbatim}
\usepackage[hyphens]{url}
\usepackage{microtype}
\begin{document}
\title{Calling to CNN-LSTM for Rumor Detection: \\A Deep Multi-channel Model for Message Veracity Classification in Microblogs}
\titlerunning{Calling to Deep CNN-LSTM for Rumor Detection}

%
\author{Abderrazek Azri \inst{1} \and C\'ecile Favre \inst{1} \and
Nouria Harbi \inst{1} \and \\J\'er\^ome Darmont  \and Camille No\^us\inst{2}}
\authorrunning{A. Azri et al.}
%
\institute{Université de Lyon, Lyon 2, UR ERIC \\ 5 avenue Pierre Mend\`es France, F69676 Bron Cedex, France \and Université de Lyon, Lyon 2, Laboratoire Cogitamus \\
\email{$\{$a.azri,cecile.favre,nouria.harbi,jerome.darmont$\}$@univ-lyon2.fr}\\
\email{camille.nous@cogitamus.fr}}
\toctitle{Calling to CNN-LSTM for Rumor Detection: \\A Deep Multi-channel Model for Message Veracity Classification in Microblogs}
\tocauthor{Abderrazek~Azri, C\'ecile~Favre, Nouria~Harbi, J\'er\^ome~ Darmont, Camille~No\^us}
\maketitle              
\setcounter{footnote}{0}
\begin{abstract}
Reputed by their low-cost, easy-access, real-time and valuable information, social media also wildly spread unverified or fake news. Rumors can notably cause severe damage on individuals and the society. Therefore, rumor detection on social media has recently attracted tremendous attention. Most rumor detection approaches focus on rumor feature analysis and social features, i.e., metadata in social media. Unfortunately, these features are data-specific and may not always be available. In contrast, post contents (including images or videos) play an important role and can indicate the diffusion purpose of a rumor. Furthermore, rumor classification is also closely related to opinion mining and sentiment analysis. Yet, to the best of our knowledge, exploiting images and sentiments is little investigated. Considering the available multimodal features from microblogs, notably, we propose in this paper an end-to-end model called deepMONITOR that is based on deep neural networks, by utilizing all three characteristics: post textual \textit{and} image contents, as well as sentiment. deepMONITOR concatenates image features with the joint text and sentiment features to produce a reliable, fused classification. We conduct extensive experiments on two large-scale, real-world datasets. The results show that deepMONITOR achieves a higher accuracy than state-of-the-art methods. 

\end{abstract}
\section{Introduction}
\label{sec:introduction}
Nowadays, more and more people consume news from social media rather than traditional news organizations, thanks to social media features such as information sharing, real time, interactivity, diversity of content and virtual identities. However, conveniently publishing news also fosters the emergence of various rumors and fake news that can spread promptly through social networks and result in serious consequences.

To detect rumors on microblogs, which we particularly target in this paper,  most existing studies focus on the social features available in social media. Such features are post metadata, including the information on how post propagate, e.g., the number of retweets, followers, hashtags (\#), user information, etc. To exploit such features, many innovative solutions \cite{castillo2011information,ruchansky2017csi}  have been proposed. Unfortunately, these features are not always available, e.g., in case the rumor has just been published and not yet propagated, and do not indicate the purpose of a rumor, which is one of its most important aspects. Moreover, although social features are useful in rumor analysis, contents reveal more relevant in expressing the diffusion purpose of rumors \cite{lin2015rumor}. Hence, in this paper, we analyse message contents from three aspects to automatically detect rumors in microblogs.

First, social media messages have rich textual contents. Therefore, understanding the semantics of a post is important for rumor detection. Attempts to automate the classification of posts as true or false usually exploit natural language processing and machine learning techniques that rely on hand-crafted and data-specific textual features \cite{castillo2011information,kwon2017rumor}.
These approaches are limited because the linguistic characteristics of fake news vary across different types of fake news, topics and media platforms.
Second, images and videos have gained popularity on microblogs recently and attract great attention. Rich visual information can also be helpful in classifying rumors \cite{jin2017multimodal}. 
Yet, taking images into account for verifying post veracity is not sufficiently explored, with only a few recent studies exploiting  
multimedia content \cite{jin2016novel,jin2017multimodal}.
Third, liars can be detected, as they tend to frequently use words carrying negative emotions out of unconscious guilt \cite{newman2003lying}. Since emotion is closely related to fake news  \cite{ajao2019sentiment}, analyzing emotions with opinion mining and sentiment analysis methods may help classifying rumors. 

Automating rumor detection with respect to one of the three characteristics mentioned above is already challenging. Hand-crafted textual features are data-specific and time consuming to produce; and linguistic characteristics are not fully understood. Image features and emotions, which are a significant indicators for fake news detection in microblogs, are still insufficiently investigated.

To address these limitations, we propose an end-to-end model called deepMONITOR,  based on deep neural network that are efficient in learning textual or visual representations and that jointly exploits textual contents, sentiment and images. To the best of our knowledge, we are the first to do this. 
Hence, deepMONITOR can leverage information from different modalities and capture the underlying dependencies between the context, emotions and visual information of a rumour.

More precisely, deepMONITOR is a multi-channel deep model where we first employ a Long-term Recurrent Convolutional Network (LRCN) to capture and represent text semantics and sentiments through emotional lexicons. This architecture combines the advantages of Convolutional Neural Network (CNN) for extracting local features and the  memory capacity of  Long Short-Term Memory Networks (LSTM) to  connect the extracted features well. Second, we employ the pretrained  VGG19  model \cite{simonyan2014very} to extract salient visual features from post images. Image  features are then fused with the joint representations of text and  sentiment to classify messages. Eventually, we experimentally show that deepMONITOR outperforms state-of-the-art rumor detection models on two large multimedia datasets collected from Twitter.

The remainder of this paper is organized as follows. In Section~\ref{sec:relatedworks}, we survey and discuss related works. 
In Section \ref{sec:framework}, we thoroughly details the deepMONITOR framework. In Section~\ref{sec:experiments}, we experimentally validate deepMONITOR with respect to the state of the art. Finally, in  Section~\ref{sec:conclusion}, we conclude this paper and hint at future research.

\section{Related Works}
\label{sec:relatedworks}
Most studies in the literature address the automatic rumor detection task as feature-based. Features can be extracted from text, social context, sentiment and even attached images. Thus, we review existing work from the following two categories: single modality-based rumor detection and multimodal-based rumor detection.

\subsection{Monomodal-based Rumor Detection}
\subsubsection{Textual features} are extracted from textual post contents. They are derived from the linguistics of a text, such as lexical and syntactic features. In the literature, there is a wide range of textual features \cite{castillo2011information,shu2017fake}. Unfortunately, linguistic patterns are highly dependent on specific events and the corresponding domain knowledge. Thus, it
is difficult to  manually design textual features for traditional machine learning-based rumor detection models. To overcome this limitation, a Recurrent Neural Network (RNN) can learn the representations of posts in time series as textual features~\cite{ma2015detect}.

\subsubsection{Social context features} represent user engagements in news on social media,  such as the number of mentions(@), hashtags(\#) and URLs~\cite{shu2017fake}. Graph structures can capture message propagation patterns \cite{wu2015false}. However, as textual features, social context features are very noisy, unstructured and require intensive labor to collect. Moreover, it is difficult to detect rumors using social context-based methods when the rumor has just popped up and not yet propagated, i.e., there is no social context information.

\subsubsection{Visual features} are typically extracted from images and videos. Very few studies address the verification of multimedia content credibility on social media. Basic message features are characterized \cite{gupta2013faking,wu2015false} and various visual features are extracted \cite{jin2016novel}. Visual features include clarity, coherence, diversity and clustering scores, as well as similarity distribution histogram. However, these features remain hand-crafted and can hardly represent complex distributions of visual contents.

\subsubsection{Sentiment features} are emotional signals. There exists a relationship between rumors and sentiments in messages and an emotion feature, i.e., the ratio of the count of negative and positive words, can be built \cite{ajao2019sentiment}. Besides, emotion features can also be extracted with respect to emotional lexicons from news contents \cite{giachanou2019leveraging}.

\subsection{Multimodal Rumor Detection} 

To learn feature representations from multiple aspects, deep neural networks, and especially CNNs and RNNs, are successfully applied to various tasks, including visual question answering \cite{antol2015vqa}, image captioning \cite{karpathy2015deep} and rumor detection \cite{jin2017multimodal,zhou2020mathsf}. In \cite{jin2017multimodal} authors propose a deep model uses attention mechanisms to fuse and capture the relations between visual features and joint textual/social features. Yet, it is very hard to identify high-level visual semantics in rumor detection, compared with object-level semantics in traditional visual recognition tasks. As a result, there is no mechanism that explicitly guarantees the learning of this matching relation in the attention model. 

Zhou et al.\cite{zhou2020mathsf} propose a neural-network-based method named SAFE that utilizes news multimodal information for fake news detection, where news representation is learned jointly by news textual and visual information along with their relationship (similarity). Assessing the similarity between text and image helps classify rumors where objects in the image are not mentioned in the text. Yet, other types of rumors escape this rule, e.g., caricatures widely used by journalists, where the text might be very different from the image, while it does not necessarily mean that the article is fake.

\section{deepMONITOR Model}
\label{sec:framework}
In this section, we formally define the problem and introduce some key notations, then introduce the components of deepMONITOR.      

\subsection{Problem Definition and Model Overview}
\begin{figure*}[htbp]
	\centerline{\includegraphics[width=0.85\textwidth]{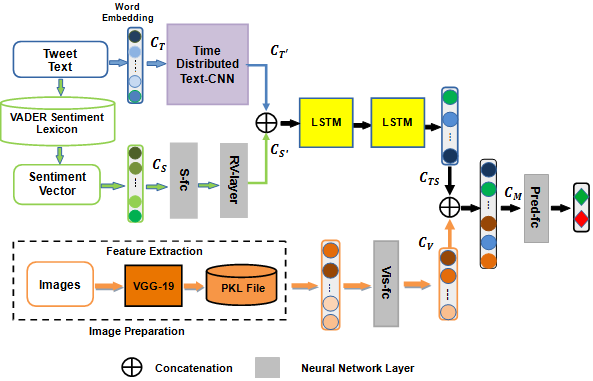}}
	\caption{Overview of deepMONITOR}
	\label{Fig1} 
\end{figure*}

We define a message instance as $M=\{T,S,V\}$ consisting of textual information $T$, Sentiment information $S$, and visual information $V$. We denote $C_T$, $C_S$ and $C_V$ the corresponding representations. Our goal is to learn a discriminable feature representation $C_M$ as the aggregation of $T$, $S$ and $V$ for a given message $M$, to predict whether $M$ is a fake ($\hat{y} = 1$) or a real message ($\hat{y} = 0$). First, we learn text with a CNN, then we merge the output with a sentiment vector with two stacked LSTMs, which generates a joint representation $C_{TS}$ for these two modalities. Visual feature $C_V$ is obtained with a pretrained deep CNN model. Finally, $C_{TS}$ and $C_V$ are concatenated to form the final multimodal feature representation $C_{M}$ of message $M$. $C_{M}$ is the input of a binary classifier that predicts whether the message instance is fake or real. A global overview of deepMONITOR is presented in Figure \ref{Fig1}.

\subsection{LSTM Networks}
\begin{figure}[htbp]
	\centerline{\includegraphics[width=0.4\textwidth]{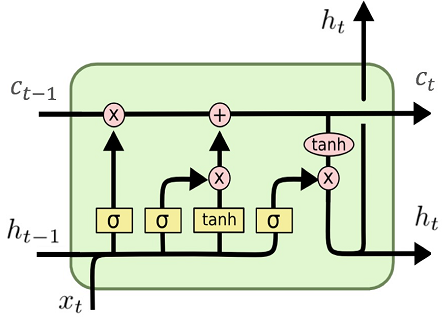}}
	\caption{Structure of an LSTM cell}
	\label{Fig2} 
\end{figure}

For completeness, we present a brief introduction of the sequential LSTM model. LSTM is a special type of feed-forward RNN that can be used to model variable-length sequential information. Its structure is shown in Figure \ref{Fig2}.

Given an input sequence $\{x_1,x_2...,x_T\}$, a basic RNN model generates the output sequence $\{y_1,y_2...,y_T\}$, where $T$ depends on the length of the input. Between the input layer and the output layer, there is a hidden layer, and the current hidden state $h_t$ is estimated using a recurrent unit: 
\begin{equation} \label{eq1}
h_t = f(h_{t-1},x_t)
\end{equation}
where $x_t$ is the current input, $h_{t-1}$ is the previous hidden state and $f$ can be an activation function or other unit accepting both $x_t$ and $h_{t-1}$ as input and producing the current output $h_t$. 

To deal with vanishing or exploding gradients \cite{bengio1994learning,pascanu2013difficulty} in
learning long-distance temporal dependencies, LSTMs extend
basic RNNs by storing information over long time periods
in elaborately designed memory units. Specifically, each LSTM cell $c$ is controlled by a group of sigmoid gates: an input gate $i$, an output gate $o$ and a forget gate $f$ that remembers the error during error propagation~\cite{hochreiter1997long}. For each time step $t$, the LSTM cell receives input from the current input $x_t$, the previous hidden state $h_{t-1}$ and the previous memory cell $c_{t-1}$. These gates are updated \cite{gers2002learning,hochreiter1997long} as follows:
\begin{equation}
i_t = \sigma(W^i_x x_t + W^i_h h_{t-1} + b_i)
\end{equation}
\begin{equation}
f_t = \sigma(W^f_x x_t + W^f_h h_{t-1} + b_f)
\end{equation}
\begin{equation}
o_t = \sigma(W^o_x x_t + W^o_h h_{t-1} + b_o)
\end{equation}
\begin{equation}
\tilde{c_t} = tanh(W^c_x x_t + W^c_h h_{t-1} + b_c)
\end{equation}
\begin{equation}
c_t = f_t \odot c_{t-1} + i_t \odot \tilde{c_t}
\end{equation}
\begin{equation}
h_t = o_t \odot tanh(c_t)
\end{equation}
where $W^i_.$, $W^f_.$, $W^o_.$ are weight matrices for corresponding gates, and $b_.$ are bias terms that are learned from the network. $\odot$ denotes the element-wise multiplication between two vectors. $\sigma$  is the logistic sigmoid function. $tanh$ is the hyperbolic tangent function. The input gate $i$ decides the degree to which new memory is added to the memory cell. The forget gate $f$ determines the degree to which the existing memory is forgotten. The memory cell $c$ is updated by forgetting part of the existing memory and adding new memory $\tilde{c}$.
\subsection{Multimodal Feature Learning}

\subsubsection{Text Feature Extraction}

\begin{figure*}[htbp]
	\centerline{\includegraphics[width=0.75\textwidth]{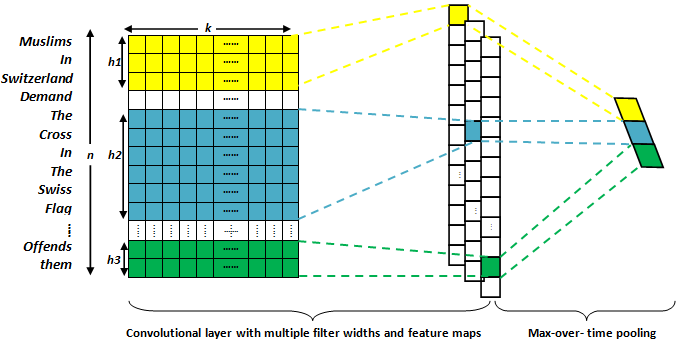}}
	\caption{Text-CNN architecture}
	\label{Fig3} 
\end{figure*}

To extract informative features from textual contents, we employ a CNN. CNNs have indeed been proven to be effective in many fields. We incorporate a modified CNN
model, namely a Text-CNN \cite{kim-2014-convolutional}, in our textual feature extraction. The architecture of the Text-CNN is shown in Figure \ref{Fig3}.

The Text-CNN takes advantage of multiple filters with various window sizes to capture different granularities of features.
Specifically, each word in the message is first represented as a word embedding vector that,  for each word, is initialized with a pretrained word embedding model. Given a piece of message with $n$ words, we denote as $T_i \in R^k$ the corresponding $k$ dimensional word embedding vector for the $i^{th}$ word in the message. Thus, the message can be represented as:
\begin{equation}
T_{1:n} = T_1 \oplus T_2 \oplus ... \oplus T_n
\end{equation}
where $\oplus$ is the concatenation operator. To produce a new feature, a convolution filter with window size $h$ takes the contiguous sequence of $h$ words in the message as input. For example, the feature $t_i$ generated from a window size $h$ starting with the $i^{th}$ word, can be represented as:
\begin{equation}
t_i = \sigma(W_c . T_{i:i+h-1} + b_c)
\end{equation}
where, $W_c \in R^{hk}$ and $b_c \in R$ are the weight and bias of the filter, respectively, and $\sigma$ is the rectified linear activation function (ReLU). This filter is applied to each possible window of $h$ words in the message to produce a feature map:
\begin{equation}
t = [t_1, t_2, ..., t_{n-h+1}]
\end{equation}

For every feature vector $t \in R^{n-h+1}$, we then apply a max-pooling operation to capture the most important information. Now, we get the corresponding feature for one particular filter. The process is repeated until we get the features of all filters. In order to extract textual features with different granularities, various window sizes are applied. For a specific window size, we have $d$ different filters. Thus, assuming there are $c$ possible window sizes, we have $c \times d$ filters in total. Following the max-pooling operations, a flatten layer is needed  to  ensure that the representation of the textual features $C_{T\prime} \in R^{c \times d}$ is fed back as input to the LSTM network.

Note that the Text-CNN above is only capable of handling a single message, transforming it from input words into an internal vector representation. We want to apply the Text-CNN model to each input message and pass on the output of each input message to the LSTM as a single time step. Thus, We need to repeat this operation across
multiple messages and allow the next layer (LSTM) to build up internal state and update weights across a sequence of the internal vector representations of input messages. Thus, we wrap each layer in the Text-CNN in a Time-Distributed layer \cite{chollet2015keras}. This layer achieves the desired outcome of applying the same layers multiple times and providing a sequence of message features to the LSTM to work on. 

\subsubsection{Sentiment Feature Extraction}

\begin{figure*}[htbp]
	\centerline{\includegraphics[width=0.75\textwidth]{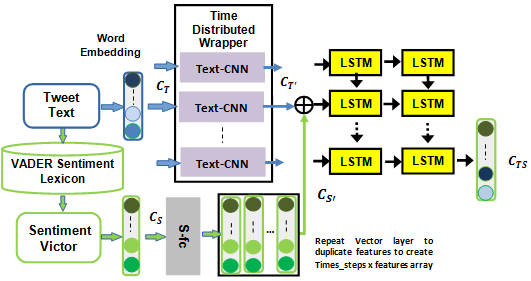}}
	\caption{Fusion process of text and sentiment features with Text-CNN and LSTM}
	\label{Fig4} 
\end{figure*}

We hypothesize that incorporating emotional signals into the rumor classification model should have some benefits. 
To extract emotional signals
from messages, we adopt a lexicon-based approach, i.e., the Valence Aware Dictionary and sEntiment Reasoner (VADER), which is a lexicon and rule-based sentiment analysis tool that is specifically attuned to sentiments expressed in social media~\cite{gilbert2014vader}.  This model  is sensitive to both the polarity (positive/negative) and the intensity (strength) of emotion. VADER relies on a dictionary that maps lexical features to emotion intensities known as sentiment scores. The sentiment score of a text can be obtained by summing up the intensity of each word in the text. In addition, we calculate some textual features that express specific  semantics  or  sentiments, such as  emotional  marks  (question  and exclamation  marks) and  emoticons. We form the initial sentiment representation $C_S = [s_1, s_2, ...,s_l]^T$, where $l$  is the dimension of sentiment features and $s_i$ is the scalar value of the $i^{th}$ dimension. We first use a fully connected layer (S-fc in Figure \ref{Fig1})  to output a proper representation of sentiment vector $C_{S\prime}$:
\begin{equation}
C_{S\prime} = W_{sf}C_S  
\end{equation}
where $W_{sf}$ are weights in the fully-connected layer. Then, we use a Repeat Vector layer \cite{chollet2016keras} to ensure that $C_{S\prime}$ has the same dimension (3D) as the representation of the textual features $C_{T\prime}$. To  connect the extracted features well, the representations of sentiment and those of textual features are then concatenated and fed as input to a two-stacked LSTM. Stacking LSTM hidden layers makes the model deeper, enables a more complex representation of our sequence data, and captures information at different scales. At each time step $i$, the LSTM takes as input $[C_{{T_i}\prime},C_{S\prime}]$, i.e., the concatenation of the $i^{th}$ message $C_{T_i\prime}$ and the transformed sentiment feature $C_{S\prime}$. The resultant joint representation of text and sentiment features, denoted as $C_{TS} \in R^p$, has the same dimension (denoted as  $p$) as the visual feature representation that is addressed in the next subsection. The whole process is illustrated in Figure \ref{Fig4}. 

\subsubsection{Image Feature Extraction}

The images attached to messages form the input of the visual sub-network (the bottom branch in Figure \ref{Fig1}). We employ the pretrained VGG-19 model \cite{simonyan2014very} to generate visual neurons as image features. We retain all front layers of the VGG-19 model and remove the last dense output layer, as well as the classification output layer.
We extract the features from all  images and store them into files. The benefit is that the very large pretrained VGG-19 does not need to be loaded, held in memory and used to process each image while training the textual submodel. For each loaded visual feature, we add a fully connected layer (Vis-fc in Figure \ref{Fig1}) to adjust the dimension of the final visual feature representation  $C_V \in R^p$, as follows:
\begin{equation}
C_V = \psi(W_{vf}C_{V_{vgg}})
\end{equation}
where $C_{V_{vgg}}$ is the visual feature representation obtained from pretrained VGG-19, $W_{vf}$ is the weight matrix of the fully connected layer and $\psi$ denotes the ReLU activation function.
The resultant joint representation of textual and sentiment features $C_{TS}$ and the visual feature representation $C_V$ are then concatenated to form the final multimodal feature representation of a given message, denoted as $C_M = C_{TS} \oplus C_{V} \in R^{2p}$.

\subsection{Model Learning}

Till now, we have obtained the joint multimodal feature representation $C_M$ of a given message $M$, which is fed into a first fully connected layer with ReLu activation function, and a second fully connected layer with sigmoid activation function to predict whether the messages are fake. The output of the sigmoid layer for the $i^{th}$ message, denoted as $p(C_{M^i})$, is the probability of this post being fake:
\begin{equation}
p(C_{M^i}) = \sigma (W_{df2}\psi(W_{df1}C_{M^i}))
\end{equation}
where $W_{df1}$ and $W_{df2}$ are weights in the two fully-connected layers, $C_{M^i}$ is the multimodal representation of the $i^{th}$ message instance and $\sigma$ and $\psi$ are the sigmoid and ReLu functions, respectively.
We employ the cross-entropy to define the detection loss of $i^{th}$ message:
\begin{equation}
L(M^i) = -y^i\log{p(C_{M^i})} - (1 - y^i)\log{(1 - p(C_{M^i}}))
\end{equation}
where $y^i$ represents the ground truth label of the $i^{th}$ message instance with 1 representing false messages and 0 representing real messages. 
To minimize the loss function, the whole model is trained end-to-end with batched Stochastic Gradient Descent:
\begin{equation}
L = -\frac{1}{N}\sum_{i=1}^{N}[y^i\log{p}(C_{M^i})+(1-y^i)\log{(1 - p(C_{M^i}}))]
\end{equation}
where $\mathnormal{N}$ is the total number of message instances. 

\section{Experimental Validation}
\label{sec:experiments}
In this section, we first detail two real-world social media datasets used in our experiments. Then, we present the state-of-the-art rumor detection approaches, followed by the details of our experimental setup. We finally analyze the performance of deepMONITOR with respect to existing methods. 

\subsection{Datasets}

To provide a fair evaluation deepMONITOR's performance, we conduct experiments on two real-world social media datasets collected from Twitter. Let us first detail both datasets.

\subsubsection{FakeNewsNet  \cite{shu2020fakenewsnet}} 
is one of the most comprehensive fake news detection benchmark. Fake and real news articles are collected from the fact-checking websites PolitiFact 
and GossipCop. 
Ground truth labels (fake or true) of news articles in both datasets are provided by human experts, which guarantees the quality of labels. We consider that all the tweets that discuss a particular news article bear the truth value, i.e., the label of the article, because it contributes to the diffusion of a rumor (true or false), even if the tweet denies or remains skeptical regarding the veracity of the rumor.

Since we are particularly interested in images in this work, we extract and exploit the image information of all tweets. We first remove duplicated and low-quality images. We also remove duplicated tweets and tweets without images, finally obtaining 207,768 tweets with 212,774 attached images. We carefully split the  training and testing datasets so that tweets concerning the same events are not contained in both the training and testing sets.

\subsubsection{DAT@Z20} 
is a novel dataset we collected from Twitter. More concretely, we retrieve all statements and reports of various nature verified by human experts from a fact-checking website; specifically contents published on June 1\textsuperscript{st}, 2020. To guarantee a high quality ground truth, we retain only the data and metadata from 8,999 news articles explicitly labeled as fake or real. To extract tweets that discuss news articles, we create queries with the most representative keywords from the articles' abstracts and titles. Then, we refine  keywords by adding, deleting or replacing words manually with respect to  each  article's context. 
We use the Twitter API to obtain the searched tweets by sending, as arguments, the queries prepared previously. Moreover, we employ Twitter Get status API to retrieve the available surrounding social context (retweets, reposts, replies, etc.) of each tweet.

Since we aim to build a multimedia dataset with images, we collect both the tweets' textual contents and attached images. Thus, from the 2,496,982 collected tweets, we remove text-only tweets and duplicated images to obtain 249,076 tweets with attached images. Finally, we split the whole dataset into  training and testing sets and ensure that they do not contain any common event. Tweets take the label of the news articles they refer to, for the same reason as above. 

The detailed statistics of the two datasets are shown in Table~\ref{tab1}.

\begin{table}
	\caption{Dataset statistics}\label{tab1}
	\centering
	\begin{tabular}{|l|c|c|c|c|c|c|}
		\hline
		\multirow{2}{*}{\textbf{Statistics \textbackslash}{\textbf{ Dataset}}}&\multicolumn{3}{c|}{\textbf{FakeNewsNet}}&\multicolumn{3}{c|}{\textbf{DAT@Z20}}\\
		\cline{2-7}
		&\textbf{True}&\textbf{Fake}&\textbf{Overall}&\textbf{True}&\textbf{Fake}&\textbf{Overall}\\
		\hline
		News articles&17,441&5,755&23,196&2,503&6,496&8,999\\
		\hline
		News articles with images&17,214&1,986&19,200&455&858&1313\\
		\hline
		All Tweets&1,042,446&565,314&1,607,760&875,205&1,621,775&2,496,982\\
		\hline
		Tweets with images&161,743&46,025&207,768&81,452&167,624&249,076\\
		\hline
		Images&163,192&49,582&212,774&93,147&202,651&295,798\\
		\hline
	\end{tabular}
\end{table}
\subsection{Experimental Settings}

To learn  a textual representation of tweets, we use the pretrained GloVe word embedding model \cite{pennington2014glove} after standard text preprocessing. We obtain a $k = 50$-dimensional word embedding vector for each word in both datasets. One reason to choose the GloVe model is that the embedding is trained on tweets. 
We set the Text-CNN network's filters number to $d = 32$ and the window size of filters to \{4, 6, 8\}. We extract 14 sentiment features from both datasets (Table \ref{tab2}). The hidden size of the fully connected layer of sentiment features is 32. 
The joint representation of text and sentiment uses a first LSTM with hidden size 64 and a second LSTM with hidden size 32. 
\begin{table}
	\caption{Sentiment features' details}\label{tab2}
	\centering
	\begin{tabular}{|l|}
		\hline
		\thead{\textbf{Feature}}\\
		\hline
		Vader Negative/Positive/Neutral/Compound Score \\
		\# positive/negative words, Fraction of positive/negative words\\
		\# sad/happy emoticons, \# exclamation/question mark \\
		\# uppercase characters, words/characters \\
		\hline
	\end{tabular}
\end{table}

Image features come from the output of the antepenultimate layer of the pretrained VGG-19 model, to generate a 4096-dimensional vector. This vector is fed to a fully connected layer with hidden size 32. The final multimodal feature representation is fed into a fully connected layer with hidden size 10. deepMONITOR uses a batch size of 64 instances. In our experiments, each dataset was separated into
70\% for training and 30\% for testing. The number of iterations is 100 in the training stage with an early stopping strategy on both datasets. The  learning rate is $10^{-2}$. 

\subsection{Baselines}
We compare deepMONITOR with three groups of baseline methods: monomodal methods, multimodal methods, and a variant of deepMONITOR.

\subsubsection{Monomodal Methods}
We propose three baselines, where text, sentiment and image information are used separately for rumor classification.
\begin{itemize}
	\item \textbf{Text}: deepMONITOR using textual information only.
	\item \textbf{Image}: deepMONITOR using visual information only.
	\item \textbf{Sent}: deepMONITOR using sentiment information only.
\end{itemize}

\subsubsection{Multimodal Methods}

We compare deepMONITOR with two state-of-the-art methods for multi-modal rumor detection.
\begin{itemize}
	\item \textbf{att-RNN} \cite{jin2017multimodal} is a deep model that employs LSTM and VGG-19 with attention  mechanism to fuse textual, visual and social-context features of news articles. We set the hyper-parameters as in \cite{jin2017multimodal} and exclude the social context features for a fair comparison.
	
	\item \textbf{SAFE} \cite{zhou2020mathsf} is a neural-network-based method that explores the relationships (similarities) between the textual and visual features in news articles. We set the hyper-parameters as in \cite{zhou2020mathsf}. 
\end{itemize}

Eventually, we also include a variant \textbf{deepMONITOR-} of deepMONITOR, where sentiment information is removed.

\subsection{Performance Analysis}

We first present the general performance of deepMONITOR by comparing it with baselines. Then, we conduct a component analysis by comparing deepMONITOR with its variants. Finally, we analyze the LRCN part. 
We use accuracy, precision, recall, and ${F_{1}}$ score as evaluation metrics.

\subsubsection{General Performance Analysis}

Table \ref{tab3} shows the experimental results of baselines and deepMONITOR on FakeNewsNet and DAT@Z20. We can observe that the overall
performance of deepMONITOR is significantly better than the baselines
in terms of accuracy, recall and ${F_{1}}$ score. 
Moreover, the general performance of multimodal methods is deepMONITOR $>$ SAFE $>$ att-RNN.  deepMONITOR indeed achieves an overall accuracy of 94.3\% on FakeNewsNet set and 92.2\% on DAT@Z20, which indicates it can learn effectively the joint features of multiple modalities. Compared to the state-of-the-art methods, deepMONITOR achieves an accuracy improvement of more than 6\% and 8\% with respect to SAFE; and 15\% and 18\% with respect to att-RNN, on FakeNewsNet and DAT@Z20, respectively. 

\begin{table}
	\caption{Performance comparison}\label{tab3}
	\centering
	\begin{tabular}{|c|c|c|c|c|c||c|c|c|}
		\hline
		\multicolumn{2}{|c|}{}&\textbf{Text}&\textbf{Image}&\textbf{Sent}&\textbf{\thead{deep\\MONITOR-}}&\textbf{\thead{att-\\RNN}}&\textbf{SAFE}&\textbf{\thead{deep\\MONITOR}}\\
		\hline
		&\textbf{Acc.}&0.865&0.776&0.650&0.874&0.799&0.888&\textbf{0.943}\\
		\cline{2-9}
		&\textbf{Prec.}&0.875&0.775&0.638&0.932&0.787&0.866&\textbf{0.934}\\
		\cline{2-9}
		\textbf{FakeNews} &\textbf{Rec.}&0.852&0.778&0.698&0.808&0.823&0.943&\textbf{0.955}\\
		\cline{2-9}
		\textbf{Net}&\textbf{$\bm{F_{1}}$}&0.863&0.777&0.667&0.865&0.805&0903&\textbf{0.944}\\
		\hline\hline
		&\textbf{Acc.}&0.840&0.714&0.568&0.885&0.742&0.842&\textbf{0.922}\\
		\cline{2-9}
		&\textbf{Prec.}&0.847&0.728&0.574&0.928&0.774&0.843&\textbf{0.938}\\
		\cline{2-9}
		\textbf{DAT@Z20} &\textbf{Rec.}&0.829&0.684&0.532&0.836&0.582&0.903&\textbf{0.905}\\
		\cline{2-9}
		&\textbf{$\bm{F_{1}}$}&0.838&0.705&0.552&0.880&0.665&0.872&\textbf{0.921}\\
		\hline
	\end{tabular}
\end{table}

\subsubsection{Component Analysis}

\begin{figure*}[htbp]
	\centering
	\subfloat[FakeNewsNet]{\includegraphics[width=0.46\textwidth]{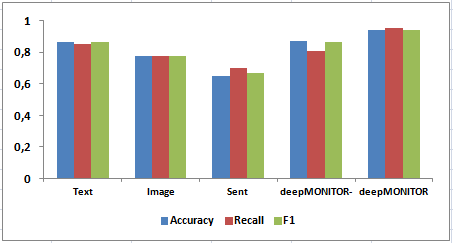}}%
	\quad
	\subfloat[DAT@Z20]{\includegraphics[width=0.46\textwidth]{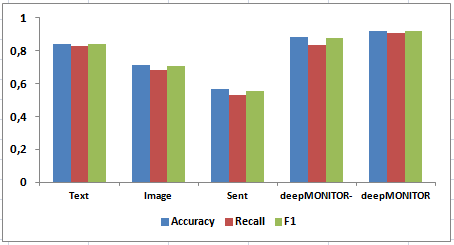}}%
	\caption{Component analysis results}%
	\label{Fig5}%
\end{figure*}

The performance of deepMONITOR and its variants are presented in Table~\ref{tab3} and Figure~\ref{Fig5}. Results hint at the following insights.

\begin{enumerate}
 \item Integrating tweets' textual information, sentiment and image information performs best among all variants. This confirms that integrating multiple modalities works better for rumor detection. 

 \item Combining textual and visual modalities (deepMONITOR-) performs better than monomodal variants because, when learning textual information, our model employs a CNN with multiple filters and different word window sizes. Since the length of each message is relatively short (smaller than 240 characters), the CNN may capture more local representative features, which are then fed to LSTM networks to deeply and well connect the extracted features. 

 \item The performance achieved with textual information is better than that of visual information. Textual features are indeed more transferable and help capture the more shareable patterns contained in texts to assess the veracity of messages. The reason is probably that both dataset have sufficient data diversity. Thus,  useful linguistic patterns can be extracted for rumor detection. 

 \item Visual information is more important than sentiment information. Although images are challenging in terms of semantics, the use of the powerful tool VGG19 allows extracting useful features representations.

 \item The performance achieved with sentiment information is the worst among multimodal variants, because without textual and visual contents, the actual meaning of tweets is lost. However, its contribution is non-negligible since the use of sentiment features (deepMONITOR- vs. deepMONITOR) can improve accuracy by 6\% and 4\%  on FakeNewsNet and DAT@Z20, respectively.
\end{enumerate}

\subsubsection{LRCN Analysis}

In this subsection, we analyze the importance of the LRCN component from the quantitative and qualitative perspectives.

\paragraph{Quantitative Analysis}

\begin{figure}[ht]
		\centering
		\subfloat{\includegraphics[width=0.5\linewidth]{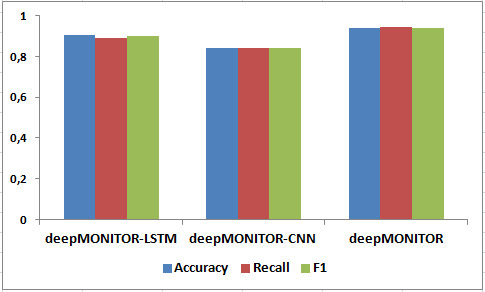}}%
		\caption{Performance comparison of the LRCN component}%
		\label{Fig6}%
\end{figure}

From deepMONITOR, we design two new models, removing the text-CNN in the first (deepMONITOR-CNN), and the two LSTM networks in the second (deepMONITOR-LSTM). Then, we run the two models on the FakeNewsNet dataset. Figure \ref{Fig6} displays the results in terms of ${F_{1}}$ score and
accuracy.  Figure \ref{Fig6}  shows that both accuracy and ${F_{1}}$ score of
deepMONITOR are better than those of deepMONITOR-CNN and deepMONITOR-LSTM. 

\paragraph{Qualitative Analysis}

\begin{figure}
\captionsetup[subfloat]{farskip=0.5pt,captionskip=0.5pt}
    	\centering
		\subfloat[deepMONITOR-LSTM]{\includegraphics[width=0.33\textwidth]{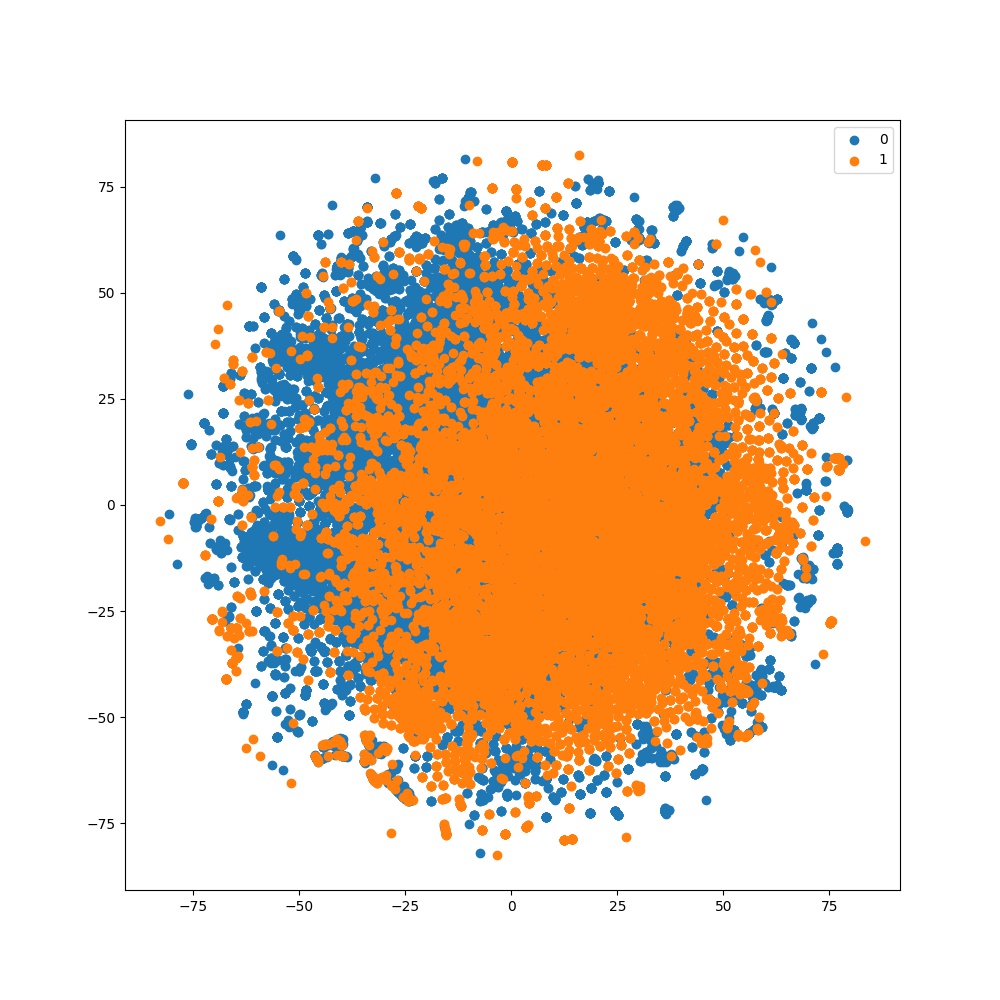}}
		\subfloat[deepMONITOR-CNN]{\includegraphics[width=0.33\textwidth]{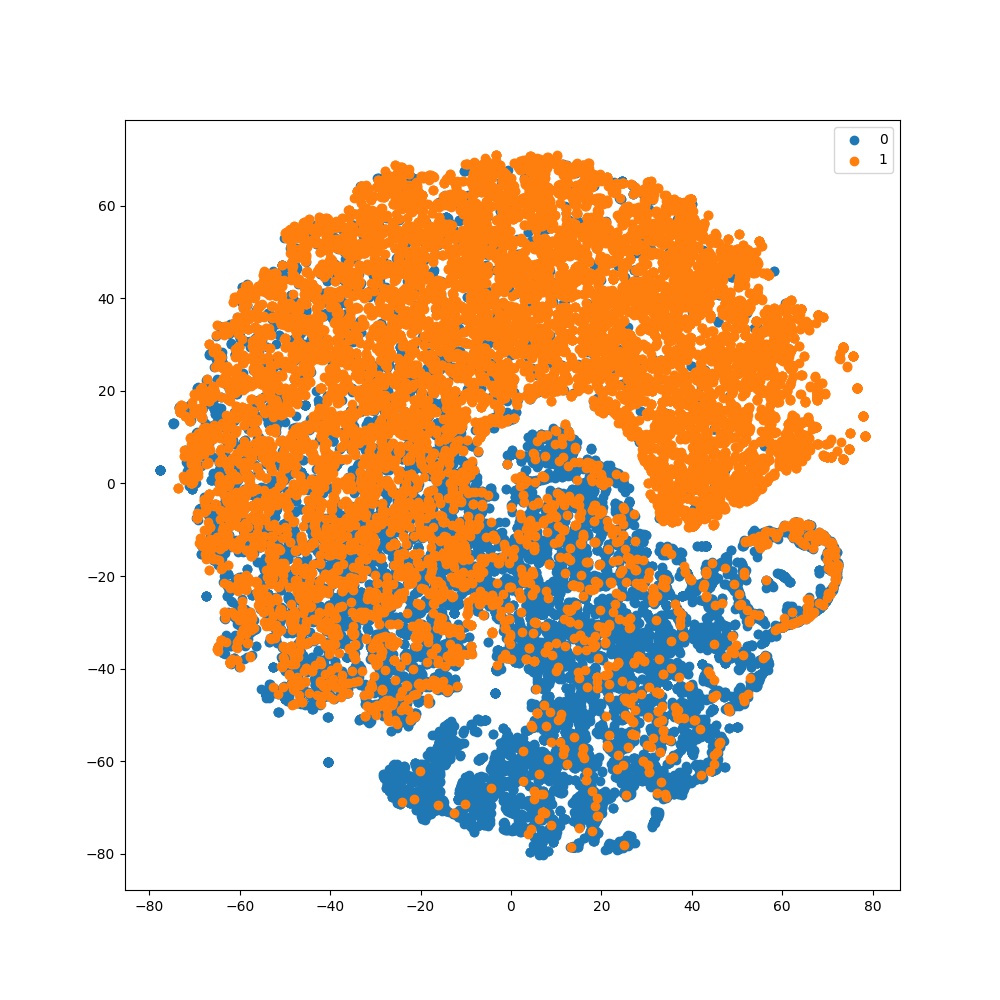}}
		\subfloat[deepMONITOR]{\includegraphics[width=0.33\textwidth]{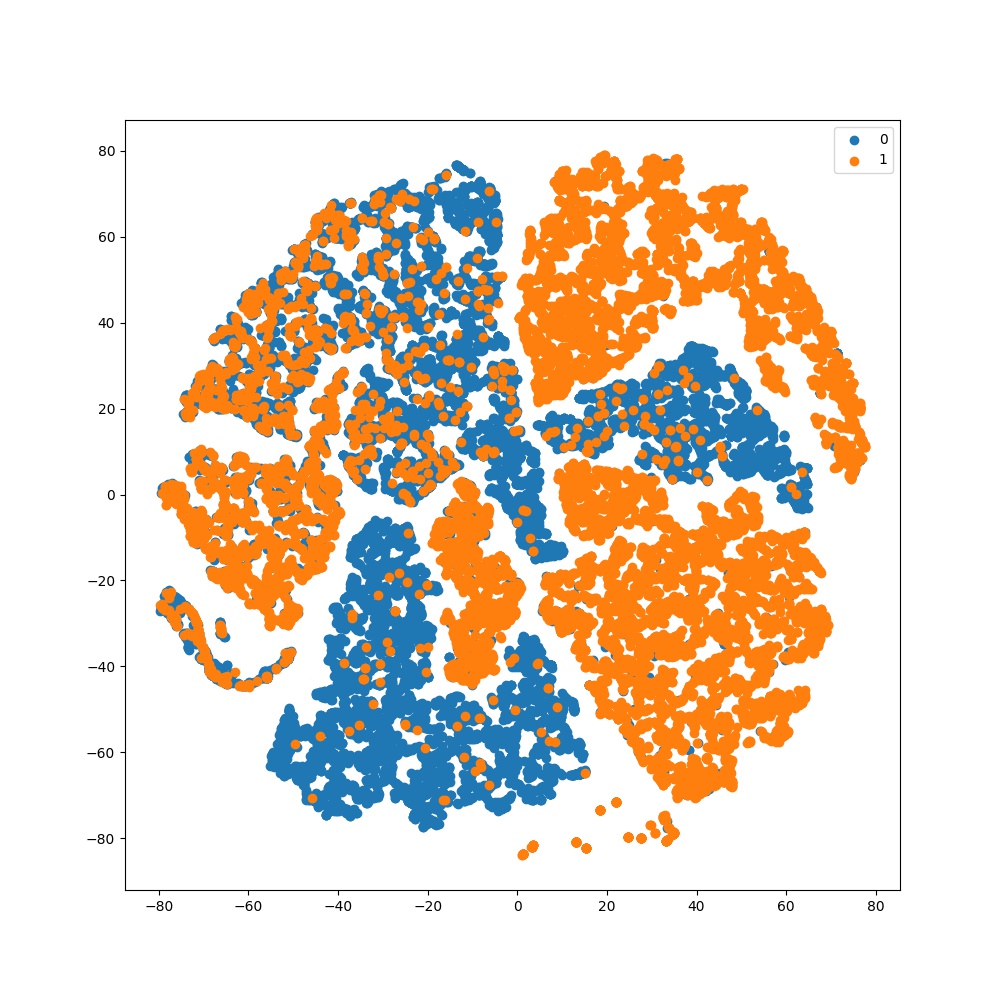}}%
		\caption{Visualizations of learned latent text and sentiment feature representations on the testing data of FakeNewsNet (the orange colored points are fake tweets and the blue ones are real)} \label{Fig7}
\end{figure}

To further analyze the importance of the LRCN component in deepMONITOR, we qualitatively visualize the feature representation $C_{TS}$ learned by deepMONITOR, deepMONITOR-CNN and deepMONITOR-LSTM on the testing data of FakeNewsNet  with t-SNE \cite{van2008visualizing} (Figure \ref{Fig7}). The label of each post is fake (orange color) or real (blue color). We can observe that deepMONITOR-CNN and deepMONITOR-LSTM can learn discriminable features, but the learned features are intertwined. In contrast, the feature representations learned by deepMONITOR are more discriminable and there are bigger segregated areas among samples with different labels. This is because, in the training stage, the Text-CNN can effectively extract local features and the LSTM networks connect and interpret the features across time steps. Thus, we can draw the conclusion that incorporating the LRCN component is essential and effective for the task of rumor detection. 

\section{Conclusion}
\label{sec:conclusion}

In this paper, we propose deepMONITOR, a deep hybrid model for rumour classification in microblogs. The model extracts and concatenates textual, visual and sentiment information altogether. For a given message, we first fuse text and emotional signals with an LRCN network, which is an appropriate architecture for problems that have a 1-dimension structure of words in a sentence, such as microblog posts. This joint representation is then fused with image features extracted from a pretrained deep CNN. Extensive experiments on two large-scale dataset collected from Twitter show that deepMONITOR outperforms
state-of-the-art methods. 

A future line of research is to further investigate the contribution of sentiment features in the detection of rumors. Dedicating a deep submodel for learning such features instead of using our current, lexicon-based approach could indeed further improve the performance of deepMONITOR.
  
%
%
%
%

%

\end{document}